%% file: paper.tex
\documentclass[letterpaper, 10 pt, conference]{ieeeconf}  

\IEEEoverridecommandlockouts                              

\usepackage{graphics} 
\usepackage{times} 
\usepackage{amsmath} 
\usepackage{amssymb}  
\usepackage{bm}

\usepackage{subcaption}
\usepackage{graphicx}
\usepackage{xcolor}
\usepackage{booktabs}
\usepackage{xspace}

\newcommand{\hide}[1]{}

\title{\LARGE \bf
Learning to Design and Construct Bridge without Blueprint
}

\author{Yunfei Li$^{1}$, Tao Kong$^{2}$, Lei Li$^{3}$, Yifeng Li$^{2}$ and Yi Wu$^{1,4}$
\thanks{$^{1}$Institute for Interdisciplinary Information Sciences, Tsinghua University, Beijing, China.
        {\tt\small \{liyf20@mails,jxwuyi@mail\}.tsinghua.edu.cn}}%
\thanks{$^{2}$ByteDance AI Lab, Beijing, China.  {\tt\small \{kongtao, liyifeng.liyf\}@bytedance.com}
}%
\thanks{$^{3}$University of California Santa Barbara. {\tt\small lilei@ucsb.edu}
Work is done while at ByteDance.}%
\thanks{$^{4}$Shanghai Qi Zhi Institute, Shanghai, China}
}

\let\oldtwocolumn\twocolumn
\renewcommand\twocolumn[1][]{%
    \oldtwocolumn[{#1}{
    \begin{center}
           \includegraphics[width=\textwidth]{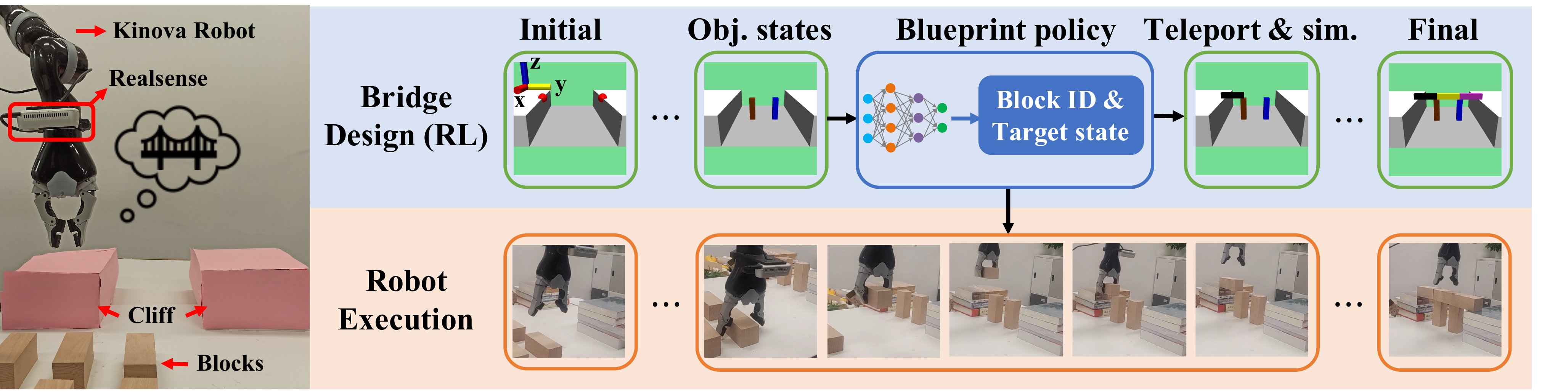}
           \captionof{figure}{Learning to design and construct bridge \textit{without a blueprint} via a bi-level approach. \textit{Top}: a neural blueprint policy learns bridge design with deep reinforcement learning in a teleportation-enabled physical simulator. \textit{Bottom}: a robot follows the pick-and-place instructions given by the blueprint policy to physically construct the bridge. }
           \label{fig:teaser}
        \end{center}
    }]
}
\begin{document}

\maketitle
\thispagestyle{empty}
\pagestyle{empty}

    
\begin{abstract}
\input{00abstract}
\end{abstract}

\IEEEpeerreviewmaketitle

\section{Introduction}

\input{10intro}  

\section{Related Work}
\input{20related}

\section{Task Setup}
\input{30setup}\label{sec:setup}

\section{Method}\label{sec:method}
\input{40method}

\section{Experiments}\label{sec:expr}
\input{50experiment}

\section{Conclusion} \label{sec:conclusion}
\input{60conclusion}


\appendix
\input{appendix}
\section*{Acknowledgments}
We would like to thank Yuchen Mo for valuable discussion about motion planning.
Lei Li is not supported by any funding in this work. 

\addtolength{\textheight}{-17.8cm}
\bibliographystyle{IEEEtran}
\bibliography{references}

\end{document}

%% file: 00abstract.tex
Autonomous assembly has been a desired functionality of many intelligent robot systems. 
We study a new challenging assembly task, designing and constructing a bridge \emph{without a blueprint}.
In this task, the robot needs to first \emph{design} a feasible bridge architecture for arbitrarily wide cliffs 
and then manipulate the blocks reliably to construct a stable bridge according to the proposed design.
In this paper, we propose a bi-level approach to tackle this task. 
At the high level, the system learns a bridge blueprint policy in a physical simulator using deep reinforcement learning and curriculum learning. 
A policy is represented as an attention-based neural network with object-centric input, which enables generalization to different number of blocks and cliff widths. 
For low-level control, we implement a motion-planning-based policy for real-robot motion control, which can be directly combined with a trained blueprint policy for real-world bridge construction  without 
tuning.
In our field study, our bi-level robot system demonstrates the capability of manipulating blocks to construct a diverse set of bridges with different architectures.

%% file: 10intro.tex

Enabling a robotic system to perform complex industrial tasks with long-term autonomy has been a long-standing research challenge. One representative benchmark of wide interest is autonomous assembly due to its mixed nature of discrete long-horizon planning and continuous motion control: the robot system needs to construct a \emph{plan}, which is typically a sequence of symbolic assembly instructions, and perform particular manipulation operations to follow the plan. Most existing assembly tasks assume a given blueprint, namely the desired configuration and relations between objects are precisely known in advance, so that the planning module only needs to figure out a physically-feasible \emph{execution order} of the predefined assembly operations from the  blueprint designed by human experts~\cite{knepper2013ikeabot}. 
However, towards a truly autonomous general-purpose robotics system, the robot should not just be capable of understanding \emph{how} to accomplish a human-decided target but also ultimately be able to deduce \emph{what} to produce to help address sophisticated real-world problems possibly beyond human capabilities.

In this paper, we consider a particularly challenging assembly problem with \emph{unknown blueprint}: \emph{ Can a robot both \emph{design} a feasible architecture and \emph{assemble} according to the design proposed by itself?} 
As a concrete example, we focus on a bridge design and construction task, where the robot is presented with two distant cliffs and a collection of blocks as construction materials. The mission is to use a minimal number of blocks to build a physically stable bridge connecting the two cliffs with the target bridge architecture unknown --- even the number of blocks to use is unspecified. This task is more difficult than standard assembly tasks with a precise goal state given. In our setting, in addition to conventional planning  for an assembly operation sequence, the robot also has to figure out a \emph{physically feasible goal state}, i.e., a stable bridge architecture, which results in an exponentially larger search space for plans.


The standard paradigm for complex assembly challenges is \emph{Task and Motion Planning} (TAMP), which decomposes a task into two components, a symbolic planning module for high-level plans and a motion planning module for execution. In classical TAMP, the planning module is purely symbolic and physics-agnostic since a physically feasible target state is typically given in a symbolic form. Hence, planning can be directly cast as a constraint-satisfying problem to which heuristic search is often applied~\cite{mcdermott1998pddl,srivastava2014combined}. There are also recent works using reinforcement learning (RL) approaches to learn a symbolic planner which can generalize better for novel assembly targets~\cite{li2020towards,zhu2020hierarchical}. However, in our setting where the blueprint, i.e., the target state, is unknown, the planning module has to be \emph{physics-aware} --- the produced sequence of assembly instructions must finally result in a physically feasible configuration, i.e., the final bridge must \emph{stably connect} the cliffs with minimal materials consumed. Therefore, physics must be considered in the planning procedure, which, however, raises a fundamental issue for classical TAMP since it can be extremely nontrivial to convert all possible physical constraints into symbolic rules for task planning. 
We additionally remark that our bridge construction task must satisfy \emph{hard constraints} for success. The final architecture will be considered feasible only if all the blocks produce a flat surface and two cliffs are stably connected --- fluctuations on the bridge surface or a tiny misplacement of blocks can result in a complete failure. Our task is much more challenging than similar assembly tasks like bin-packing, where a partial score can be achieved even if the bin is not fully packed~\cite{zhao2020online,agarwal21jampacker}.


We propose a bi-level framework to solve the bridge design and construction task: we learn a high-level blueprint policy to generate a valid construction plan towards a feasible bridge, and implements a low-level manipulation policy for executing the high-level plan. Our method is conceptually similar to standard TAMP with a major difference that the high-level blueprint policy is learned in a \emph{physics-aware} manner. Concretely, we 
use deep reinforcement learning to learn a neural blueprint policy 
in a \emph{teleportation-enabled} physical simulator. In our modified simulator, the blueprint policy does not perform any motion actions. Instead, it directly outputs a primitive instruction by selecting a particular block and specifying its target placement coordinates. Then the simulator will immediately change the internal state of the selected block so that this block will be \emph{teleported} to the desired location. Note that it is possible that such a teleportation results in collisions or other physical interactions. So, our simulator will continue physical simulation until all the blocks become stable before asking the blueprint policy for the next instruction based on the final physically-stable state. With this simulator-level modification, the blueprint policy can focus on high-level planning and also be aware of physical consequences of its plan. Moreover, to ensure that the blueprint policy can \emph{generalize} to arbitrarily wide cliffs, we adopt an attention-based policy architecture with object-centric input; to tackle the hard constraint issue, we utilize curriculum learning  as well as a very recent deep RL algorithm, Phasic Policy Gradient~\cite{cobbe20ppg}, for effective training. After the blueprint policy is learned, we can combine it with any motion-planning-based low-level policy w.r.t. the robot hardware for zero-shot deployment in the real world.



In our experiment, the trained blueprint policy can design diverse bridge architectures using different number of blocks to connect cliffs of various distances. The best trained policy achieves an 87\% training success rate for design long bridges that requires 7 blocks. We also empirically observe interesting self-correction behaviors, e.g., the learned policy can correct early block misplacement or even remove a redundant block to reduce material consumption. Ablation studies show that phasic training and curriculum learning are critical to efficient learning. Finally, we integrate the blueprint policy into a real robot system and successfully construct the designed bridges for different cliff widths.


Our contributions are summarized as follows:
\begin{enumerate}
    \item We propose a novel assembly task, bridge design and construction without blueprint, as a benchmark for intelligent robot systems;
    \item We developed a physics-aware RL framework with curriculum learning for solving this challenging bridge design and construction task;
    \item We empirically validate our approach by combining the blueprint policy with a real-world robot arm to construct bridges for various cliff widths.
\end{enumerate}

%% file: 20related.tex
Task and Motion Planning (TAMP) ~\cite{kaelbling2011hierarchical} is a standard framework to solve complex long-horizon robotic tasks, such as assembly~\cite{knepper2013ikeabot,nagele20lego}, object rearrangement~\cite{srivastava2014combined,garrett2015ffrob} and stacking~\cite{toussaint2015logic}. 
Classical TAMP methods combine symbolic planning~\cite{fahlman1974planning,mcdermott1998pddl} for sub-task decomposition and continuous-control methods for real robot execution~\cite{wolfe2010combined}. 
Our approach does not require an explicit formulation of symbolic constraints or known dynamics for plans but utilizes a model-free RL-based approach to learn to generate task plans in a physical simulator.

Many recent works utilize RL-based planners for robotic assembly and manipulation~\cite{luo19assembly,zakka2020form2fit,zhu2020hierarchical}. 
These works typically assume known goal states, either in the form of a state or an image of the target object configuration~\cite{li2020towards,zakka2020form2fit}. We focus on a more difficult setting with an unknown blueprint, which requires the agent to design the final state by itself. 
There are also related works on designing diverse structures w.r.t. a particular constraint~\cite{ritchie2015generating,ritchie2015controlling}, which is orthogonal to our focus.
Incorporating structural inductive bias into neural networks for better generalization in object-oriented tasks has attracted much research interest \cite{battaglia2016interaction,janner2018reasoning}. For example, graph neural networks and attention-based models are popular architectures for generalization over a varying number of objects~\cite{mu2020refactoring,li2020towards}.

Curriculum learning~\cite{bengio2009curriculum} is a common practice in many robotic learning tasks~\cite{akkaya2019solving,mehta2020active} by always training the agent with tasks of moderate difficulty. 
We adopt a simple curriculum similar to \cite{li2020towards} by increasing the task complexity w.r.t. the learning progress of the agent.
There are also related works on automating curriculum generation~\cite{florensa2018automatic,florensa2017reverse,sukhbaatar2018intrinsic} or developing better RL algorithms for compositional problems~\cite{haarnoja2018composable,peng2019mcp,li2021solving}, which are orthogonal to our focus.



%% file: 30setup.tex
A simulated scene of the bridge design and construction task built with MuJoCo~\cite{Todorov12mujoco} physics engine is illustrated in the top row of Fig.~\ref{fig:teaser}. The coordinating system is visualized in the ``initial'' grid, with $x, y, z$-axis rendered by red, yellow and blue, respectively.
There are two white cliffs on the left and right side of the scene. The positions of cliffs are randomly sampled at the beginning of each episode. There are $N$ cuboid building blocks positioned outside the valley region (thus invisible in the scene), and the task is to control a robot to manipulate these blocks to build a structure that connects the two cliffs.
Task completion is determined by whether there exists a flat surface above the line segment connecting the two red spots on the edge of two cliffs.
Concretely, we sample multiple points over the valley and cast a ray along the negative direction of the $z$-axis to detect the height of the construction at each point. 
If the heights at all the sampled points on the upper surface exceed the cliff height plus the thickness of the building blocks, the structure is considered a feasible bridge.
We also impose additional constraints for a smooth upper surface and fewer building blocks (more in Sec.~\ref{sec:method:formulation}).
In the real world, we use a 6-DoF Kinova Gen2 Robot Jaco2 with a parallel gripper as our robot platform (see the leftmost part of Fig.~\ref{fig:teaser}). Multiple cuboid blocks with size 5cm $\times$ 5cm $\times$ 12cm are aligned on the table in front of the cliffs. 

%% file: 40method.tex
Directly learning a flat policy to solve the entire bridge design and construction task can be remarkably difficult due to the extremely long task horizon and the exponentially large design space of possible bridge architectures. 
Therefore, we tackle this problem by decomposing it into a hierarchy of two sub-problems: a high-level problem of learning an abstract-level blueprint policy to generate assembly instructions of moving one particular building block to a desired location at one time, and a low-level problem that aims to control the robot to follow the high-level instructions. 

We adopt deep reinforcement learning for blueprint generation and standard robotic-control algorithms to solve the low-level execution problem.  
The blueprint policy is learned in a \emph{teleportation-enabled} physical simulator, where the actual execution of the robot motion is omitted. The simulator will directly teleport the selected block to the desired location and continue running the physical simulator until a stable state is reached, which not only simplifies policy learning but also ensures the policy is physics-aware.
We remark that the blueprint design problem is more difficult than classical symbolic planning problems because the learned policy also needs to be aware of all the physical constraints to produce a \emph{physically feasible} construction plan.




\input{41abstract_level}

\input{42motion_planning}

%% file: 41abstract_level.tex


\subsection{Problem Formulation of Blueprint Design}\label{sec:method:formulation}

The blueprint policy aims to sequentially generate pick-and-place instructions that can finally build a flat bridge connecting the two given cliffs with the minimal number of building blocks. In each time, the agent can observe the current scene and instruct one object to be re-configured to a new state. After running the physics engine for a fixed number of steps, the agent can receive feedback from the environment, observe the consecutive scene and give the next instruction. This procedure can be formulated as a Markov decision process (MDP) defined with the tuple $\{\mathcal{S}, \mathcal{A}, \Gamma, \mathcal{R}, T\}$, where $\mathcal{S}$ denotes the state space, $\mathcal{A}$ means the action space, $\Gamma$ is the transition function, $\mathcal{R}$ stands for the reward function and $T$ is the horizon of an episode. Each component in the tuple is detailed as follows:

\paragraph{State space} 
We use an object-centric representation to encode the states of all $N$ building blocks and 2 cliffs: $\bm{s}=[s_1, s_2, \cdots, s_{N+2}]$.
$s_i$ is a 14-D vector consisting of 3-D position, 3-D Euler angle, 3-D Cartesian velocity, 3-D angular velocity, 1-D object type indicator denoting whether the object is a building block or not, and 1-D time.

For the building blocks located outside the valley between cliffs, we treat their states as reset states and use special tokens (all ``-1''s or all ``0''s) to fill the first 12-D dimensions of $s_i$ corresponding to object configurations. In other words, the actual configurations of objects in the reset states are invisible to the RL agent.

Notice that at the beginning of each episode, all the building blocks are in reset states. Therefore it does not matter what their actual configurations are as long as they are outside the valley.

\paragraph{Action space} 
For simplicity, we only generate pick-and-place instructions that put building blocks on the $yz$-plane across the middle of the two cliffs. An action is a 4-D vector consisting of 1-D target object ID, 1-D target $y$ position, 1-D target $z$ position, and 1-D rotation angle around the $x$-axis.  
The actions that convert an object into reset states are interpreted by the simulator as special ``reset'' actions, and will directly throw the object back to its initial position. 

\paragraph{Reward function} 
The reward function is a combination of ``construction reward'', ``smoothness reward'' and ``material saving reward'' to encourage building a smooth bridge that connects the two cliffs with minimal consumption of building blocks.

\textbf{Construction reward:} 
As is described in Sec.~\ref{sec:setup}, we use ray-casting to detect whether the bridge is built. 
Suppose we sample $S$ points on the upper surface of the current structure and detect their heights as $h_i, 1 \leq i \leq S$. 
The agent will receive a big bonus $R_{succ}$ if all the heights exceeds the detecting threshold $H$, where $R_{succ} = \prod_{i=1}^{S}\mathbb{I}(h_i > H)$ and $H$ is defined as the height of the cliff plus the thickness of one building block. 
This reward may be too sparse to guide an efficient learning process, so we also credit the agent for partially built structures with a construction reward 
\begin{equation*}
    R_{cons} = \frac{\sum_{i=1}^{S} \mathbb{I}(h_i > H)}{S}.
\end{equation*}

\textbf{Smoothness reward:} 
We expect a flat upper surface of the built bridge. So we reward the agent with 
\begin{equation*}
    R_{flat} = \mathbb{I}(R_{cons} = 1) * max(d - \sum_{i=1}^{S-1} |h_{i+1} - h_i|, 0)
\end{equation*}
only after the bridge has been built. $d$ is a hyper-parameter controlling how much fluctuation we can tolerate. 

\textbf{Material saving reward: } 
We also want to encourage the agent to make full use of given materials. Therefore we define a material saving reward 
\begin{equation*}
    R_{mat} = \mathbb{I}(R_{cons} = 1) * (1 - \frac{\#\textrm{block between cliffs}}{N})
\end{equation*}
to encourage completing the bridge with as few building blocks as possible.

Putting them together, we define the total reward as a linear combination of all the components: 
\begin{equation*}
    R = c_{cons}R_{cons} + c_{succ}R_{succ} + c_{flat}R_{flat} + c_{mat}R_{mat}.
\end{equation*}
Note that $R_{flat}$ and $R_{mat}$ are active only when a bridge has been successfully constructed. 

\paragraph{Transition function}
A physical simulator accepts the instruction from the blueprint policy, then directly teleports the selected object to the instructed state and continues running physical simulation for multiple steps until the environment reaches a stable state. The simulator then returns the resulting state to the blueprint agent.
Therefore, the agent can still be informed of long-horizon physical consequences of the instructions and be incentivized to seek physically stable solutions even though we do not explicitly build the concept of physics via symbolic rules or any known dynamics models. 

\subsection{Blueprint Policy Learning}
We then detail the adopted deep reinforcement learning algorithm for tackling the aforementioned MDP. An RL agent aims to find a policy $\pi$ that maximizes the accumulated discounted reward $\pi^\star = \arg\max_{\pi}\mathbb{E}_{s, a\sim \Gamma, \pi}[\sum_{t=1}^{T}\gamma^{t}R(s_t, a_t)]$, where $\gamma$ is the discount factor. Specifically, we use a Transformer~\cite{Vaswani17attention}-based feature extractor to incorporate inductive bias on objects and relations into the policy, and PPG algorithm to efficiently train the policy.

\subsubsection{Network Architecture}
\begin{figure}[tb]
    \centering
    \includegraphics[width=0.7\linewidth]{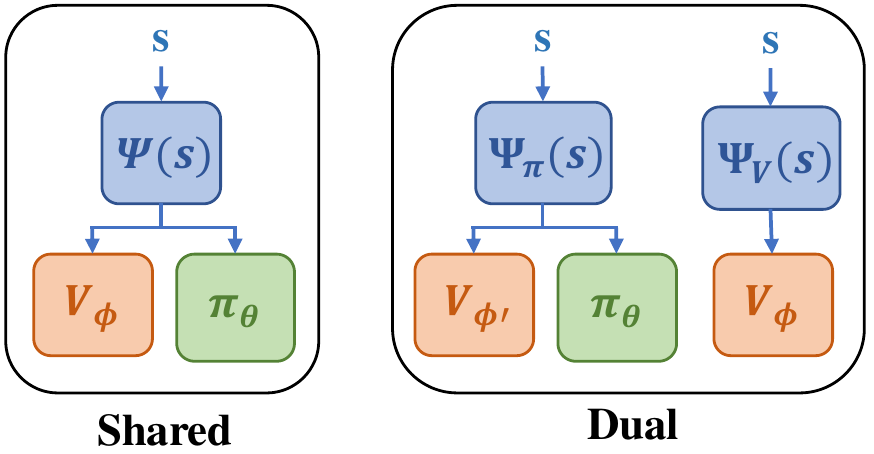}
    \includegraphics[width=0.95\linewidth]{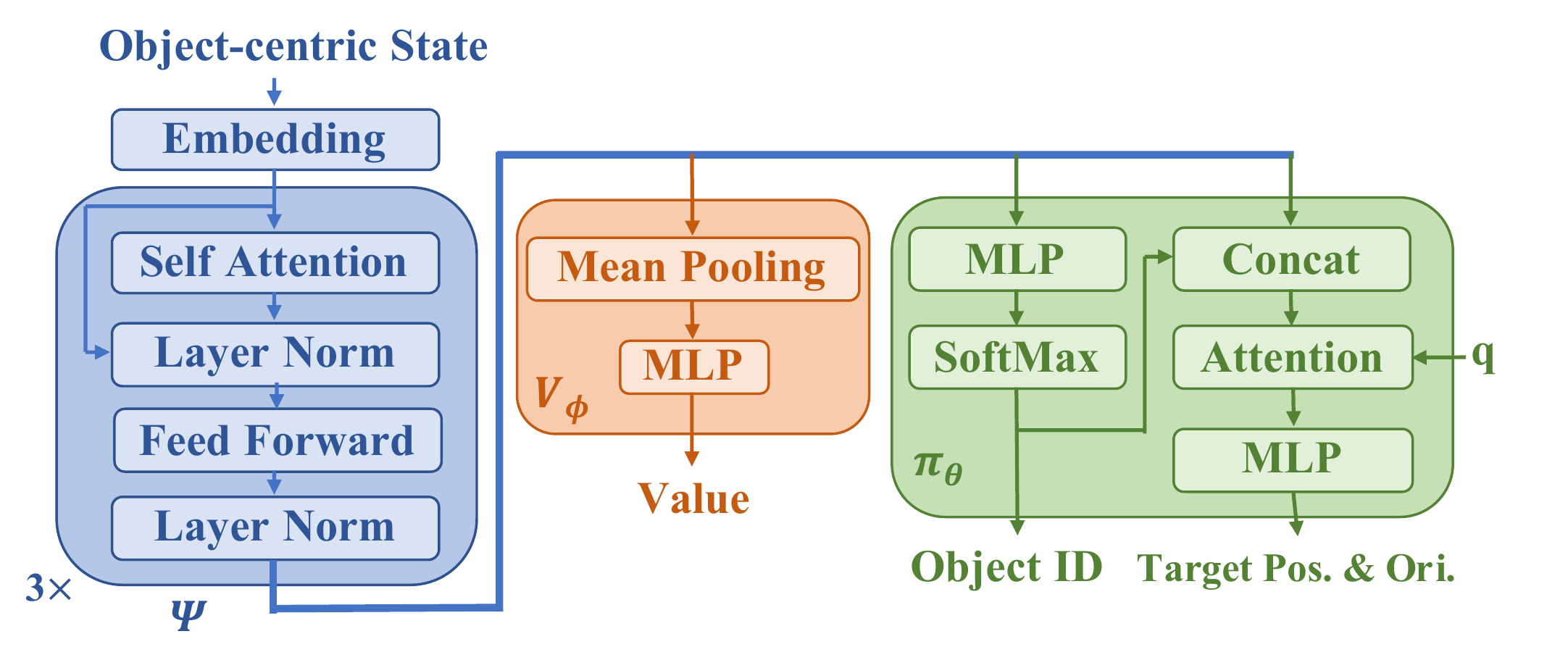}
    \caption{An overview of  blueprint policy architecture. \emph{Top}: the two structure variants, the ``shared'' and the ``dual'' network. \emph{Bottom}: the detailed attention-based network architecture.}
    \label{fig:method:arch}
\end{figure}
The overall network architecture is depicted in Fig.~\ref{fig:method:arch}.
We use an object-centric feature extractor to incorporate structural inductive bias for better generalization over objects. 
The input observations are first packed into a tensor of size $(N+2, \textrm{object\_dim})$, then converted into embeddings $x$ with a two-layer MLP. 
The embeddings are then processed with a stack of 3 attention blocks, which is similar to the Transformer encoder~\cite{Vaswani17attention}. 
Specifically, we calculate $q$, $k$ and $v$ with linear layers from embedding $x$, 
then apply a scaled dot-product self-attention function to output a feature $f = g(\frac{qk^T}{\sqrt{d}}v)$,
where $g$ is a linear layer and $d$ is the dimension of $k$. 
After a residual connection and a feed-forward network, we feed $f$ into the next attention block.
The extracted object-level feature $f$ of size $(N+2, \textrm{feature\_dim})$ will be passed into value head and policy head to get the final output. 
In the value head, $f$ is first averaged over the first dimension and is then fed into a two-layer MLP to get the value prediction. 
The policy head consists of two branches to predict target object ID and target state respectively. For the first branch, $f$ is converted to a vector of size $N+2$ with a linear layer, then normalized with softmax to get the probability mass of each object. For the second branch, $f$ is concatenated with the normalized probability from the output of the first branch, then be reduced to a single vector $f_a$ of size $\textrm{feature\_dim}$ with an attention function with a trainable query $q$. The target position and orientation are sampled from discrete distributions as suggested in~\cite{hsu2020revisiting}, and probabilities are computed by feeding $f_a$ into MLPs. 

\subsubsection{Phasic Policy Gradient (PPG) Training}
We use PPG~\cite{cobbe20ppg}, a successor of the popular on-policy RL algorithm PPO~\cite{Schulman17ppo} for policy training.
PPG adopts a phasic training process to distill value information into policy for better representation learning and uses an imitation learning objective to stabilize policy training. 
PPG suggests two architecture variants, ``dual'' and ``shared'' (see Fig.~\ref{fig:method:arch}), and we empirically find that ``shared'' performs better (more in Sec.~\ref{sec:expr:blueprint}).
In ``shared'' architecture, the policy and value network share the same feature extractor $\psi$ and later split into a policy head $\pi_{\theta}$ and a value head $V_{\phi}$. Policy and value training are separated into different phases. During the policy phase, $\psi$ and $\pi_{\theta}$ are trained with the original PPO loss $L^{\pi}$:
\begin{equation*}
    L^{\pi} = \mathbb{E}[\textrm{min}(\rho \hat{A}, \textrm{clip}(\rho, 1-\epsilon, 1+\epsilon)\hat{A})],
\end{equation*}
where $\rho = \frac{\pi_{\theta}(a_t|\psi(s_t))}{\pi_{\theta_{old}}(a_t|\psi(s_t))}$ and $\hat{A}$ is the advantage function.
In the value phase, $V_{\phi}$ is trained jointly with the policy network using $L^{joint}$, which adds a behavior cloning loss to the original value loss to force the policy output to be unchanged:
\begin{equation*}
    \begin{split}
        & L^{joint}_{\phi} = L^{V}_{\phi} + \beta_{clone}\mathbb{E}[KL[\pi_{\theta_{old}}(\cdot|\psi(s_t)), \pi_{\theta}(\cdot|\psi(s_t))]], \\
        & L^{V}_{\phi} = \frac{1}{2} \mathbb{E}[(V_{\phi}(\psi(s_t)) - V^{targ})^2].
    \end{split}
\end{equation*}
$\theta_{old}$ is the policy parameter right before the value phase.  
Therefore, the value phase distills information from value training into the shared representation and meanwhile avoids pulling the policy distribution away.
``Dual'' architecture uses separate feature extractors $\psi_{\pi}$ and $\psi_V$ for policy and value networks but attaches an auxiliary head $V_{\phi'}$ after $\psi_{\pi}$. 
In the policy phase, policy and value networks are trained together with PPO loss $L^{\pi} + L^{V}_{\phi'}$ since there are no shared parameters between them. During the value phase, $V_{\phi'}$ is trained jointly with the policy network using $L^{joint}_{\phi}$ to distill useful information into the policy representation $\psi_{\pi}$, and the value network is allowed additional training with the value loss $L^{V}_{\phi'}$.

\subsubsection{Curriculum Learning}\label{sec:method:cl} 
Since directly learning to design a long bridge is extremely difficult, we apply an adaptive curriculum that adjusts the valley width according to the training progress of the agent.
We define the hard case as when the distance between cliffs is greater than $2.5\times \textrm{block\_length}$. When an episode resets, the distance is sampled from hard cases with probability $p$ and is uniformly sampled from the whole task space $[0.5\times \textrm{block\_length}, 3.5\times \textrm{block\_length}]$ otherwise.
The curriculum starts from $p=0.01$. During training, we increase $p$ by 0.1 when the average success rate is higher than 0.6, and decrease $p$ by 0.1 when the average success rate is lower than 0.3. 

%% file: 42motion_planning.tex
\subsection{Low-Level Policy for Motion Execution}
After training a blueprint policy for producing assembly instructions, the low-level motion execution policy follows these instructions to physically manipulate the selected object to the target state. Since our blueprint policy is physics-aware during training, it can always produce physically feasible instructions for the low-level controller. Therefore, the low-level policy only needs to finish a simple pick-and-place task at each time, which can be solved by classical motion-planning algorithms. 
We generate grasp poses to the block's center of mass, and plan a collision-free path with bidirectional RRT algorithm. 
Note that since instruction generation and motion execution are fully decoupled in our approach, a learned blueprint policy can be directly applied to any real robot platform in a zero-shot manner.



%% file: 50experiment.tex

\subsection{Experiments of Blueprint Policy Learning}\label{sec:expr:blueprint}
We experiment with PPG training with shared architecture on bridge design and construction tasks with up to 7 blocks.
The performance is measured by the average success rate on the hard cases that require building a long bridge. All the experiments are repeated over 3 random seeds. All the hyper-parameters can be found in Appendix~\ref{sec:app}.
\subsubsection{Ablation Study}
\begin{figure}[tb]
    \centering
    \includegraphics[width=0.9\linewidth]{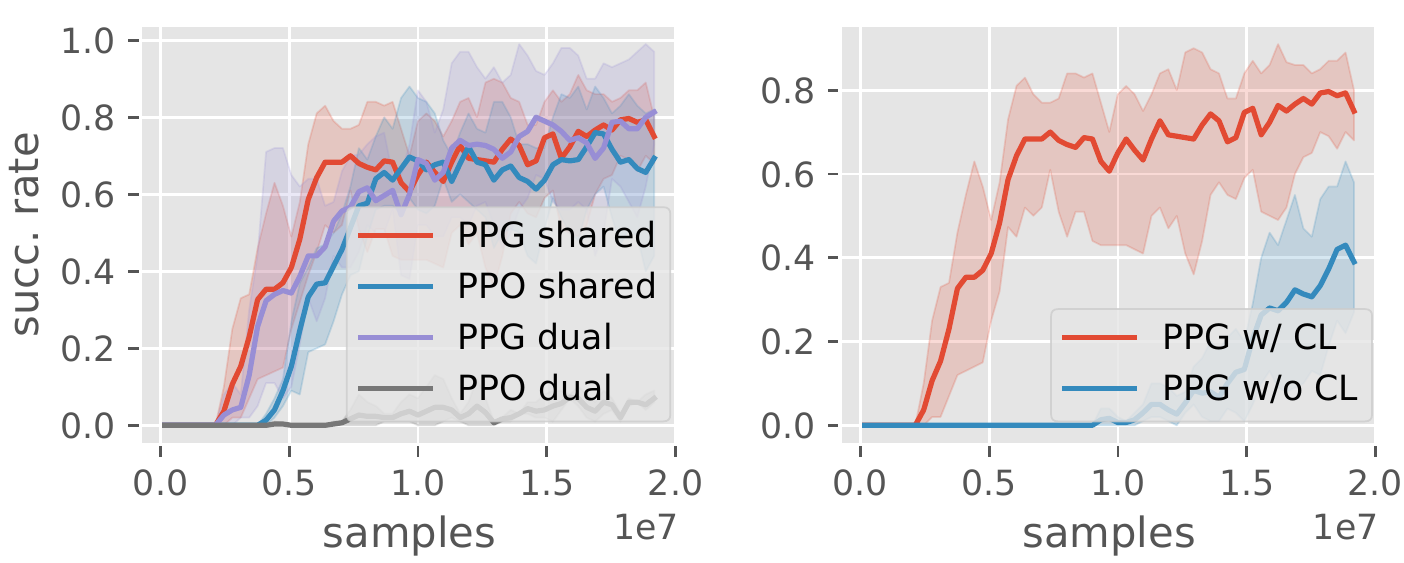}
    \caption{Ablation studies for blueprint policy learning. We evaluate training success rates on hard cases for all the variants. Our final method is in red. \emph{Left}: using PPO or PPG with different policy architectures, i.e., dual or shared network. \emph{Right}: the effectiveness of curriculum learning. 
    }
    \label{fig:expr:curve}
\end{figure}

To validate the design of our training algorithm, we compare ``PPG shared'' with the following variants as shown in the left half of Fig.~\ref{fig:expr:curve}: (a) ``PPO shared'' which replaces PPG with PPO (blue), (b) ``PPG dual'' which uses a separate feature extractor to learn value function (purple), and (c) ``PPO dual'' (grey). ``PPG shared'' generally performs the best, followed by ``PPG dual'', ``PPO shared'' and ``PPO dual''.
``PPO shared'' is significantly more sample efficient than ``PPO dual'', which verifies that sharing information in representation learning is critical for efficient policy learning.
With PPG algorithm, the sample efficiency of shared architecture can be further improved (see red and blue curves), which indicates the effectiveness of phasic training for distilling value information into policy representation. 
Although ``PPG dual'' achieves comparable performance with ``PPG shared'', its number of parameters is almost twice as many as the shared architecture due to separate feature extractors. Therefore among all the choices of algorithms and architectures, ``PPG shared'' is the best and is adopted as our final method.

We also compare our algorithm with ``PPG w/o CL'', which removes the adaptive curriculum and simply samples hard tasks with a fixed probability of 0.91. As shown in the right plot of Fig.~\ref{fig:expr:curve}, ``PPG w/o CL'' hardly makes any progress in the first 10M steps and only achieves a low success rate when ``PPG w/ CL'' has converged, which verifies that curriculum learning is critical to efficiently solving complex tasks with multiple building blocks by starting small.

We evaluate the final success rate of all the variants trained at 2e7 timesteps on 100 randomly sampled hard tasks defined in Sec.~\ref{sec:method:cl} and report their mean and standard deviation in Table.~\ref{tab:expr:evaluation}. Our algorithm ``PPG shared'' reliably solves the hard tasks and converges to an average success rate of 80.3\%. 

\begin{figure*}[tb]
\begin{minipage}{1\textwidth}
    \centering
    \begin{subfigure}[b]{0.27\textwidth}
        \centering
        \includegraphics[width=0.47\textwidth]{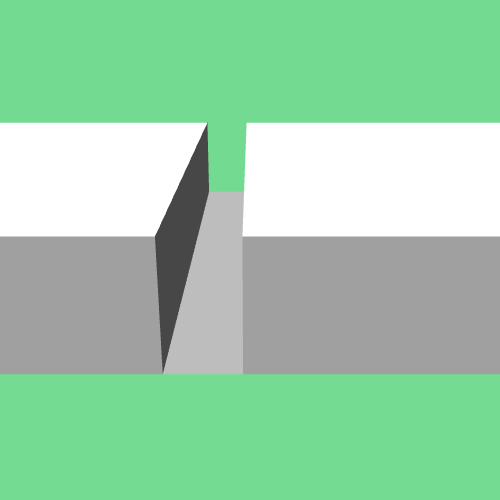}
        \includegraphics[width=0.47\textwidth]{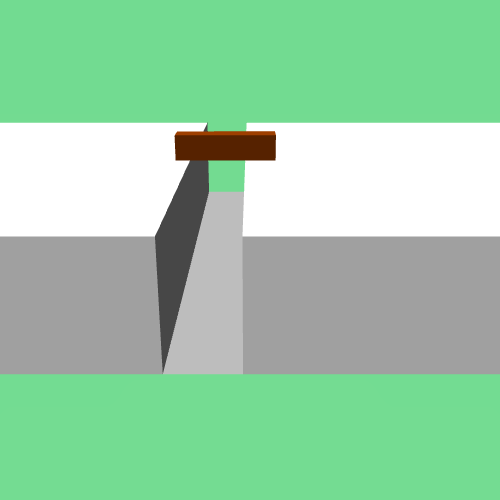}
        \caption{A short bridge with 1 block.}
        \label{fig:expr:short_bridge}
    \end{subfigure}
    \hspace{1mm}
    \begin{subfigure}[b]{0.67\textwidth}
        \centering
        \includegraphics[width=0.19\linewidth]{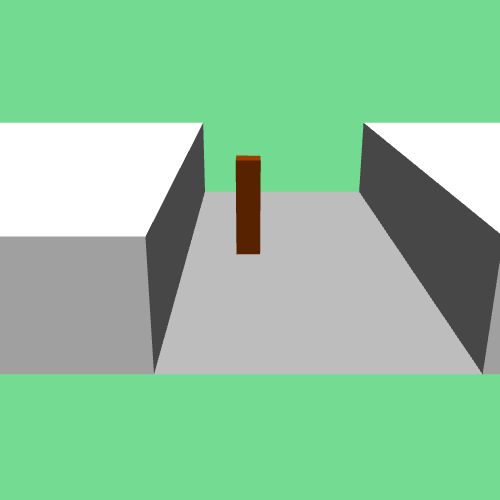}
        \includegraphics[width=0.19\linewidth]{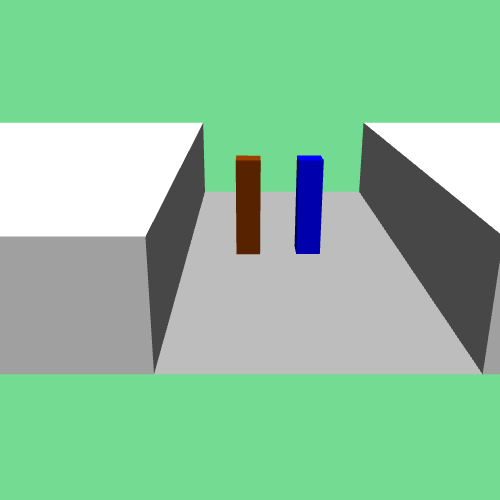}
        \includegraphics[width=0.19\linewidth]{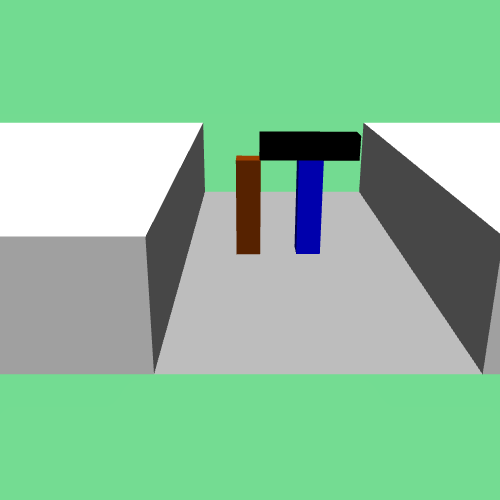}
        \includegraphics[width=0.19\linewidth]{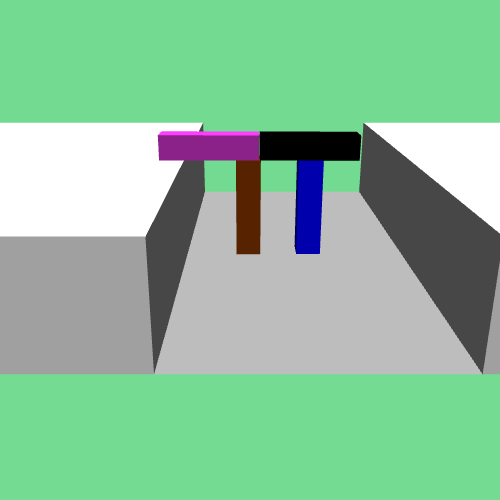}
        \includegraphics[width=0.19\linewidth]{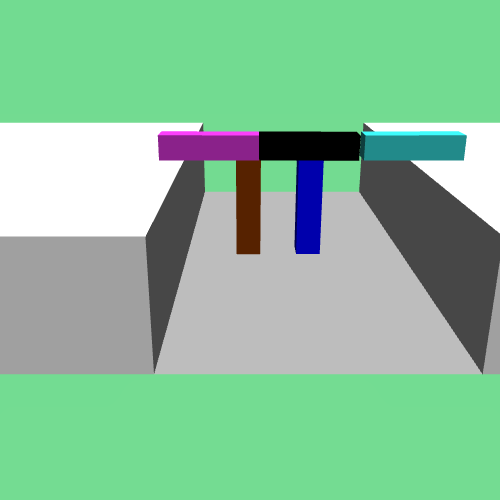}
        \caption{A medium-sized bridge with 5 blocks.}
        \label{fig:expr:medium_bridge}
    \end{subfigure}
    \begin{subfigure}[b]{1\textwidth}
        \centering
        \includegraphics[width=0.13\textwidth]{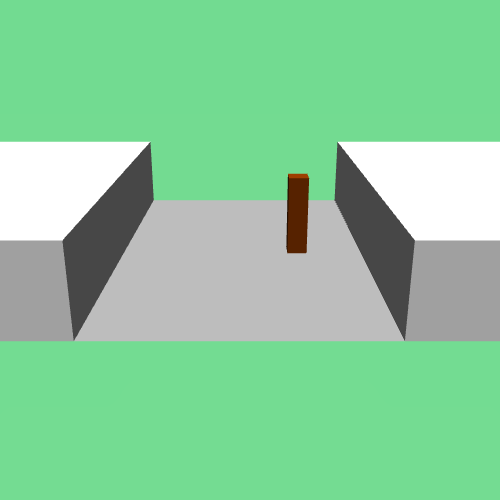}
        \includegraphics[width=0.13\textwidth]{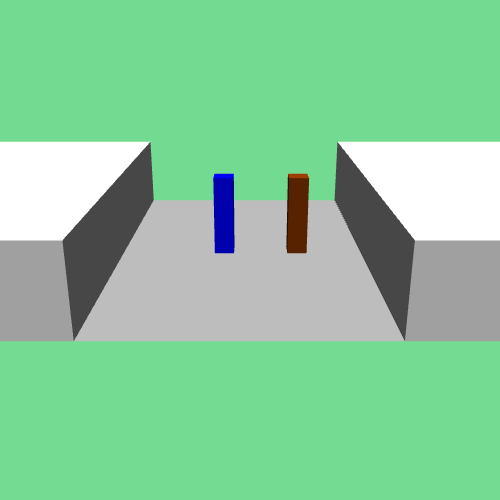}
        \includegraphics[width=0.13\textwidth]{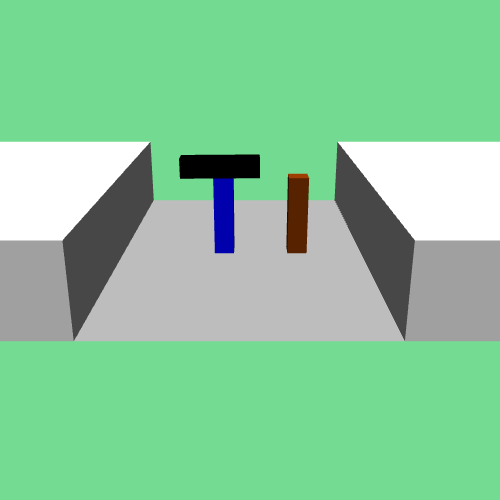}
        \includegraphics[width=0.13\textwidth]{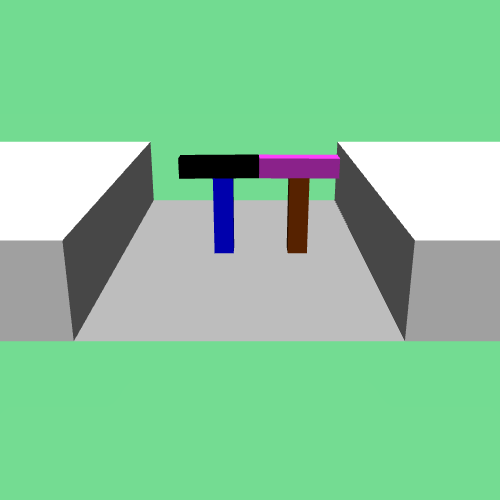}
        \includegraphics[width=0.13\textwidth]{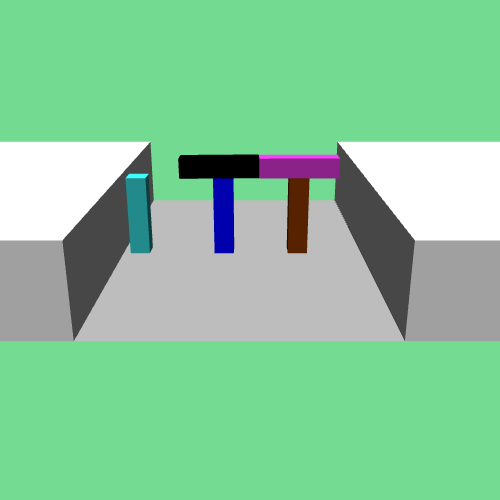}
        \includegraphics[width=0.13\textwidth]{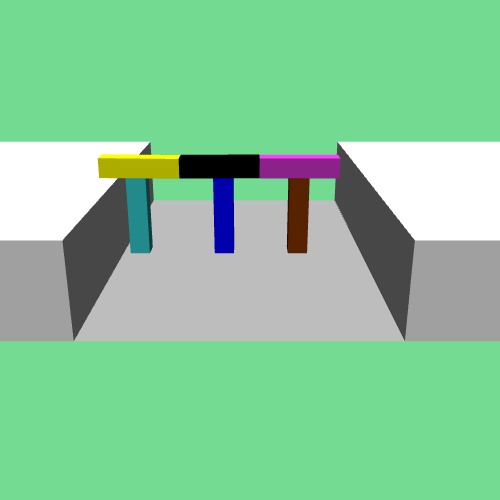}
        \includegraphics[width=0.13\textwidth]{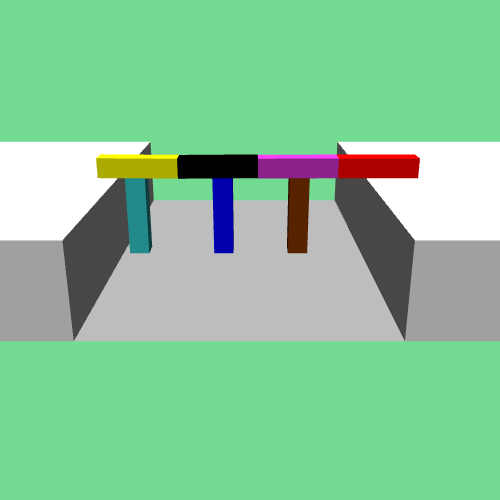}
        \caption{The longest bridge built up with 7 blocks.}
        \label{fig:expr:direct_build}
        \end{subfigure}
    
    \caption{Different bridge architectures generated by our blueprint policy when the distance between the cliffs varies.}
\end{minipage}

\begin{minipage}{1\textwidth}
    \centering
    \begin{subfigure}[b]{0.48\textwidth}
        \centering
        \includegraphics[width=0.23\textwidth]{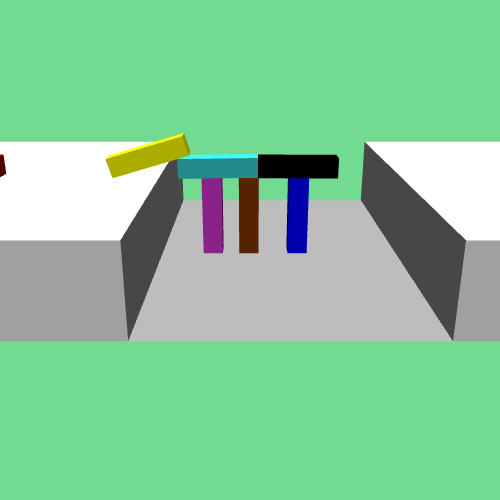}
        \includegraphics[width=0.23\textwidth]{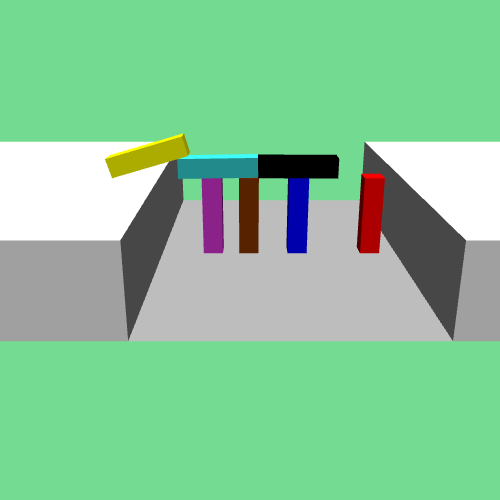}
        \includegraphics[width=0.23\textwidth]{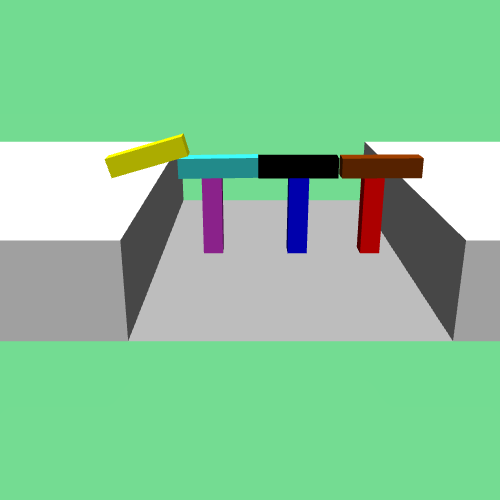}
        \includegraphics[width=0.23\textwidth]{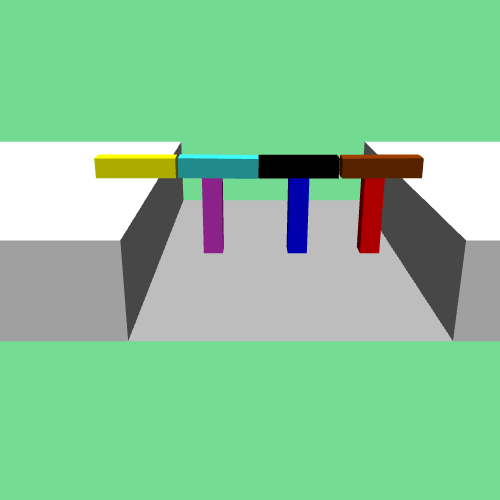}
        \caption{Adjust the blocks placed improperly (yellow) for a flat surface.}
        \label{fig:expr:build_adjust}
    \end{subfigure}
    \begin{subfigure}[b]{0.48\textwidth}
        \centering
        \includegraphics[width=0.23\textwidth]{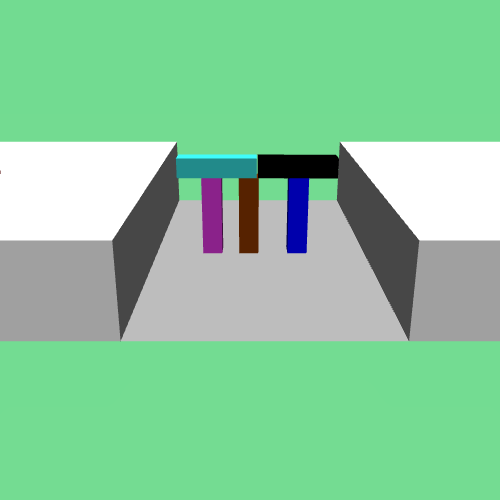}
        \includegraphics[width=0.23\textwidth]{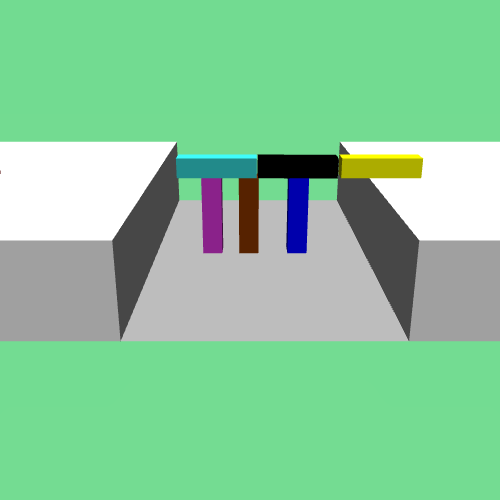}
        \includegraphics[width=0.23\textwidth]{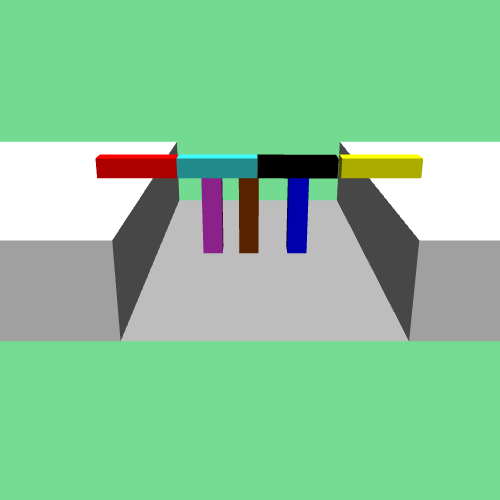}
        \includegraphics[width=0.23\textwidth]{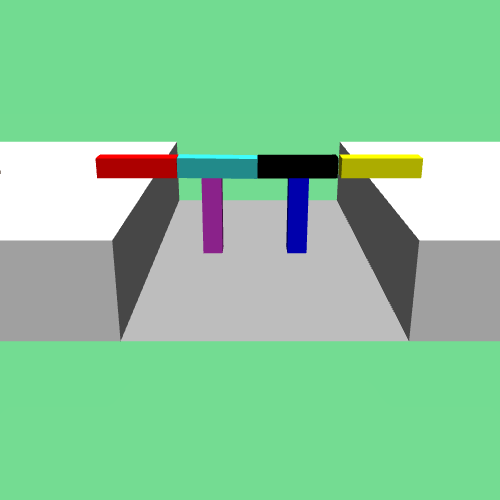}
        \caption{Remove a redundant block (brown) for fewer materials.}
        \label{fig:expr:build_remove}
    \end{subfigure}
    \caption{Emergent blueprint generation behavior. The trained policy can adjust block misplacement in the early stage.}
\end{minipage}

\end{figure*}

\begin{figure*}
    \centering
    \begin{subfigure}[b]{0.39\textwidth}
        \centering
        \includegraphics[width=0.32\linewidth]{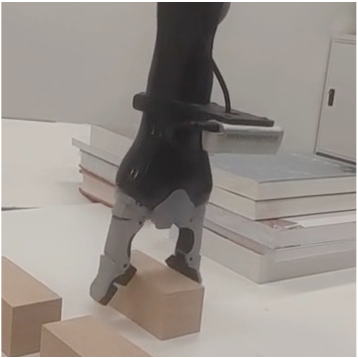}
        \includegraphics[width=0.32\linewidth]{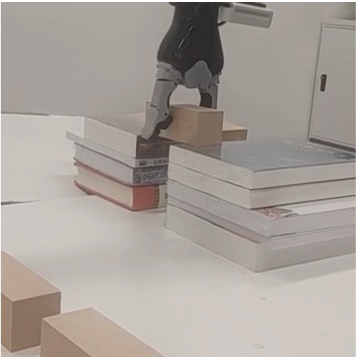}
        \includegraphics[width=0.32\linewidth]{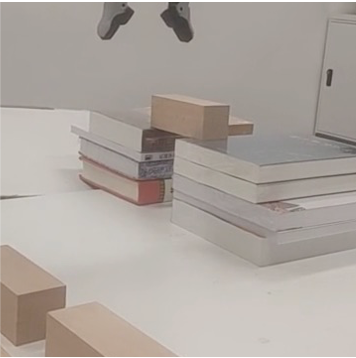}
        \caption{The simplest bridge with 1 block.}
    \end{subfigure}
    \hspace{0.5mm}
    \begin{subfigure}[b]{0.55\textwidth}
        \centering
        \includegraphics[width=0.23\linewidth]{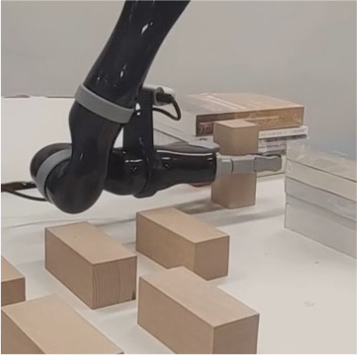}
        \includegraphics[width=0.23\linewidth]{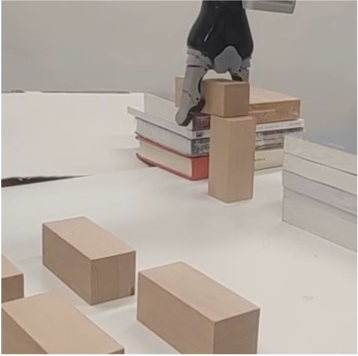}
        \includegraphics[width=0.23\linewidth]{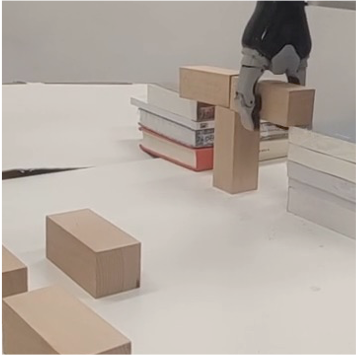}
        \includegraphics[width=0.23\linewidth]{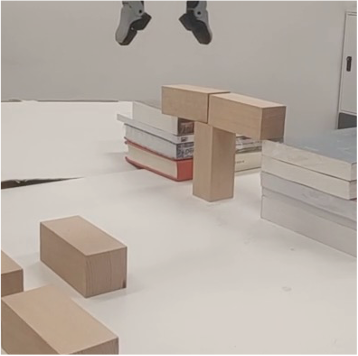}
        \caption{A T-shape bridge with 3 blocks.}
    \end{subfigure}
    \begin{subfigure}[b]{1\textwidth}
        \centering
        \includegraphics[width=0.13\linewidth]{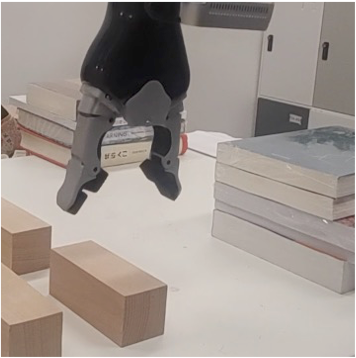}
        \includegraphics[width=0.13\linewidth]{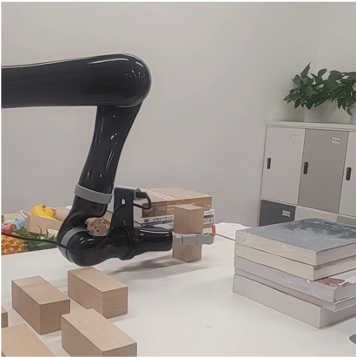}
        \includegraphics[width=0.13\linewidth]{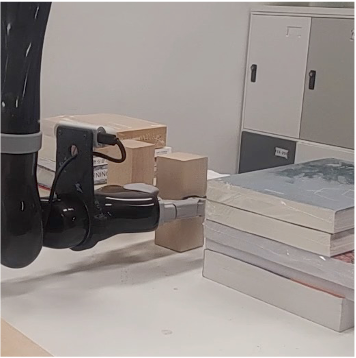}
        \includegraphics[width=0.13\linewidth]{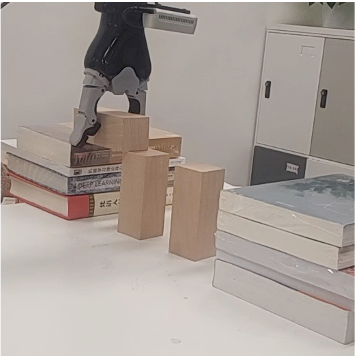}
        \includegraphics[width=0.13\linewidth]{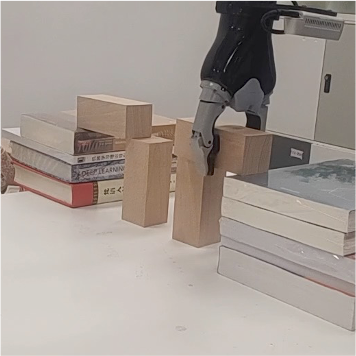}
        \includegraphics[width=0.13\linewidth]{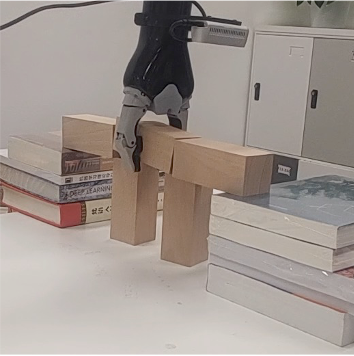}
        \includegraphics[width=0.13\linewidth]{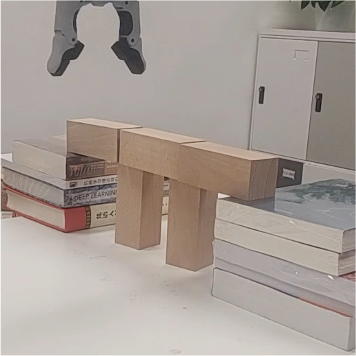}
        \caption{A $\Pi$-shape bridge with 5 blocks.}
    \end{subfigure}
    \caption{Bridge design and construction in the real world with a robot arm for different cliff widths.}
    \label{fig:expr:robot_5block}
\end{figure*}

\begin{table}[tb]
    \centering
    \caption{Mean and standard deviation of the final success rate over 3 runs evaluated on 100 hard cases.}
    \begin{tabular}{ll}
    \toprule
        Algorithm & Success rate  \\
    \midrule
        PPG shared & \textbf{0.803$\pm$0.091} \\
        PPO shared & 0.747$\pm$0.067 \\
        PPG dual & 0.803$\pm$0.184 \\
        PPO dual & 0.067$\pm$0.050 \\
        PPG shared w/o CL & 0.430$\pm$0.220 \\
    \bottomrule
    \end{tabular}
    \label{tab:expr:evaluation}
\end{table}

\subsubsection{Learned Strategies}
We then visualize the learned strategies of the blueprint RL agent. Fig.~\ref{fig:expr:direct_build} shows how the agent designs a long bridge that requires using all 7 blocks. The agent learns to first put some blocks vertically inside the valley as supporting blocks, then add more blocks horizontally on top of them to successfully connect two cliffs. 
Fig.~\ref{fig:expr:short_bridge} and Fig.~\ref{fig:expr:medium_bridge} show other modalities of discovered solutions when the valley width is shorter. The agent can use 1 block to build a shortest bridge or 5 blocks to build a medium-sized bridge. 
 The agent also demonstrates some interesting emergent behaviors that can strategically adjust the blocks improperly placed in the early stage. As shown in Fig.~\ref{fig:expr:build_adjust}, the agent adjusts the orientation of the yellow block to make a flat upper surface. In Fig.~\ref{fig:expr:build_remove}, the agent initially puts the brown block vertically as a supporting block, but then finds it redundant for the whole bridge thus removes it to save building materials.

\subsection{Real Robot Experiments}
Finally, we integrate the learned blueprint policy with a real robot system. 
\input{52robot_expr}
Fig.~\ref{fig:expr:robot_5block} shows the results of bridge design and construction deployed in the real world. We test three cases where the distances between cliffs are set to 10cm, 22cm and 32cm. The robot can successfully follow the instructions given by the learned blueprint policy to construct bridges with different modes using different number of blocks.

%% file: 52robot_expr.tex
We adopt an off-the-shelf motion planning method to control a real robot to follow the pick-and-place instructions commanded by the blueprint policy. 
We use a commercial RGB-D camera, Realsense D435i mounted on the robot wrist as vision input. A 3D oriented bounding box detection method based on principal components analysis is applied to calculate the grasp center~\cite{Zhou2018}. Then we use a top-down 6-DoF grasping algorithm 
to complete the task. 


%% file: 60conclusion.tex
In this work, we take an initial step towards robotic construction tasks without a blueprint, which requires a robot to first design the feasible structures by reasoning over the given building materials and then manipulate them with physical execution. We solve this challenging problem by decomposing it into a bi-level problem: at the high level, a blueprint policy gives pick-and-place instructions to finally achieve a feasible design; at the low level, a robot arm simply follows the given manipulation instructions. A deep reinforcement learning approach is applied to learn the blueprint policy and traditional motion-planning algorithms are applied to manipulate the objects.
We focus on building bridges in a 2D plane with identical cuboids as building blocks in the current work. In the future, we would like to extend to more complicated bridges using materials with arbitrary size and shape, with the help of more advanced exploration techniques. 

%% file: appendix.tex
\subsection{Implementation Details}\label{sec:app}
All the hyperparameters in our environment and algorithm are listed in Table.~\ref{tab:expr:hyper_algo}.
\begin{table}[!h]
    \centering
    \caption{Hyperparameters in blueprint policy learning.}
    \begin{tabular}{llll}
         \toprule
       Name & Value & Name & Value\\
    \midrule
        \#workers & 32 & T & 30 \\
        n\_steps & 1024 & $\gamma$ & 0.97\\
        n\_minibatches & 32 & $N$ & 7 \\
        n\_epochs & 10 & $d$ & 0.1\\
        entropy\_coef & 0.01 & $c_{succ}$ & 0.1 \\
        GAE-$\lambda$ & 0.95 & $c_{cons}$ & 0.05\\
        $\epsilon$ & 0.2 & $c_{flat}$ & 1.5\\
        learning rate & 2.5e-4 & $c_{mat}$ & 0.1\\
        $\beta_{clone}$ & 3 & n\_attention & 3\\
        feature\_dim & 64 & & \\
    \bottomrule    
    \end{tabular}
    
    \label{tab:expr:hyper_algo}
\end{table}

\subsection{Additional Results}
\begin{figure}[!h]
    \centering
    \includegraphics[width=0.9\linewidth]{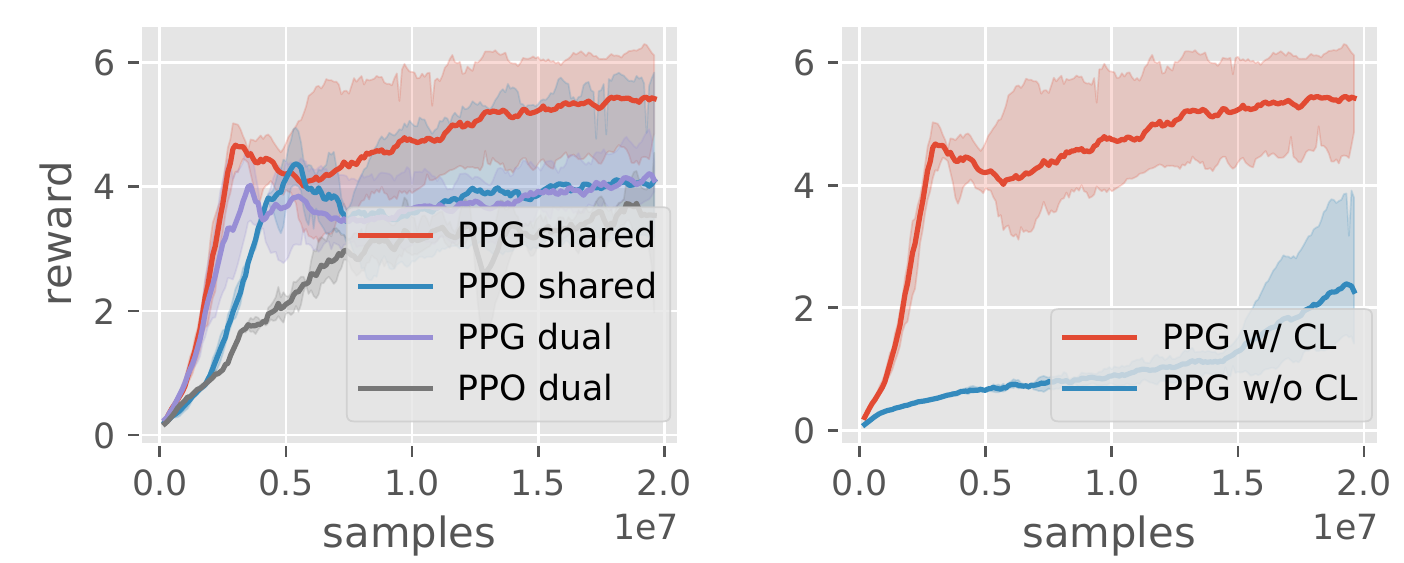}
    \caption{The mean episode reward during training time with different variants of blueprint policy learning.}
    \label{fig:app:reward_plot}
\end{figure}

We also include the curves of mean episode reward during the training process in Fig~\ref{fig:app:reward_plot}. The left plot shows the training results of different policy optimization algorithms and architectures, and the right plot shows the effect of curriculum learning. Each curve represents results from three repeated experiments.
